\documentclass[letterpaper, 10 pt, conference, final]{ieeeconf}

\IEEEoverridecommandlockouts
\pubid{979--8--3315--0964--4/24/\$31.00˜\copyright˜2024 IEEE
}

\usepackage{epsfig}
\usepackage{amsmath}
\usepackage{amssymb}
\usepackage{graphicx}
\usepackage{subfig}
\usepackage{hyperref}
\usepackage{orcidlink}

\DeclareMathOperator*{\argmin}{arg\,min}

\makeatletter
\newcommand*\titleheader[1]{\gdef\@titleheader{#1}}
\AtBeginDocument{
  \let\st@red@title\@title
  \def\@title{
    \bgroup\normalfont\centering\footnotesize\@titleheader\par\egroup
    \vskip1.2em\st@red@title}
}
\makeatother

\title{GARField: Addressing the visual Sim-to-Real gap in garment manipulation with mesh-attached radiance fields}
\titleheader{IEEE INTERNATIONAL CONFERENCE ON ROBOTICS AND BIOMIMETICS. PREPRINT VERSION. ACCEPTED OCTOBER, 2024}

\author{Donatien~Delehelle$^{\dag}$\orcidlink{0009-0000-0049-6585},~\IEEEmembership{Member,~IEEE,} Darwin~Caldwell\orcidlink{0000-0002-6233-9961},~\IEEEmembership{Fellow,~IEEE,} Fei~Chen\orcidlink{0000-0003-4397-0931},~\IEEEmembership{Senior~Member,~IEEE}%
\thanks{Manuscript received: August 31, 2024; Revised: September 2, 2024; Accepted: October 9, 2024.}
\thanks{This work is supported by IIT}
\thanks{Donatien Delehelle is with University of Genova}
\thanks{Donatien Delehelle and Darwin Caldwell are with Istituto Italiano di Tecnologia (email: {\tt\small donatien.delehelle@iit.it}, {\tt\small darwin.caldwell@iit.it})}
\thanks{Fei Chen is with the Department of Mechanical and Automation Engineering, T- Stone Robotics Institute, The Chinese University of Hong Kong, Hong Kong (e-mail: {\tt\small f.chen@ieee.org})}
\thanks{$^{\dag}$Corresponding author}
\thanks{Digital Object Identifier (DOI): 10.1109/ROBIO64047.2024.10907327}
}

\begin{document}

\maketitle
\thispagestyle{empty}
\pagestyle{empty}

\begin{abstract}

While humans intuitively manipulate garments and other textile items swiftly and accurately, it is a significant challenge for robots. A factor crucial to human performance is the ability to imagine, a priori, the intended result of the manipulation intents and hence develop predictions on the garment pose. That ability allows us to plan from highly obstructed states, adapt our plans as we collect more information and react swiftly to unforeseen circumstances. Conversely, robots struggle to establish such intuitions and form tight links between plans and observations. We can partly attribute this to the high cost of obtaining densely labelled data for textile manipulation, both in quality and quantity. The problem of data collection is a long-standing issue in data-based approaches to garment manipulation. As of today, generating high-quality and labelled garment manipulation data is mainly attempted through advanced data capture procedures that create simplified state estimations from real-world observations. However, this work proposes a novel approach to the problem by generating real-world observations from object states. To achieve this, we present GARField (Garment Attached Radiance Field), the first differentiable rendering architecture, to our knowledge, for data generation from simulated states stored as triangle meshes. Code is available on \href{https://ddonatien.github.io/garfield-website/}{the project website}.

\end{abstract}

\section{INTRODUCTION}

\PARstart{C}{lothes} and textile objects are omnipresent in everyday life. However, their manipulation is still a long-standing challenge in robotics, which hinders the development of domestic assistive robots that could drastically improve the quality of life of many people in areas such as nursing services, elderly care and hospital care.\\
\begin{figure}[ht]
	\centerline{\includegraphics[width=0.74\linewidth]{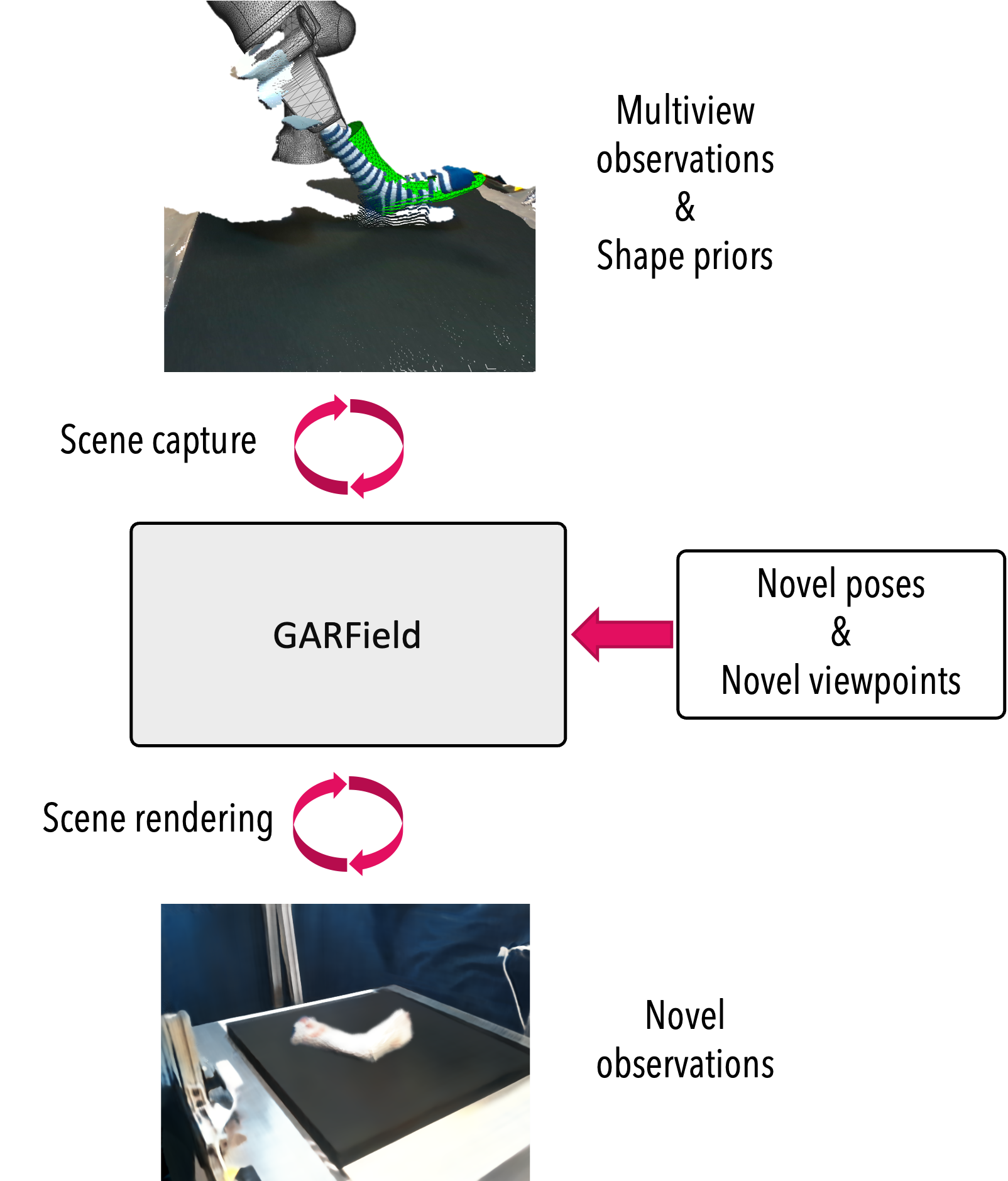}}
	\caption{\small By defining garment-attached signed distance and radiance fields, GARFIeld enables novel-view synthesis of re-posed meshes.}
	\label{overview}
\end{figure}
The critical challenges of garment manipulation stem from the non-conformity of these objects with fundamental assumptions in robotics such as rigidity, known dynamics models, and low dimensional state space \cite{longhini2024unfoldingliteraturereviewrobotic}. In the absence of such assumptions, state or partial-state estimation becomes an extremely high-dimensional prediction problem posed in partial observability conditions.\\
To circumvent this problem, some methods have abandoned state estimation and rely on task-specific extrinsic features. Wrinkles detection, for instance, has proven efficient for simple tasks such as cloth flattening \cite{sun_heuristic_based}, \cite{seita2019deeptransferlearningpick}. These features can be homogeneously defined over the whole surface of the object as they are not dependent on specific object-attached points,  which allows perception systems to locate them from purely geometrical information, such as point clouds or depth images, without occlusion reasoning. Furthermore, by default, they can be generalised to a wide range of objects of varying appearances. However, the absence of a notion of object state prevents their use in more general manipulation problems where understanding the object's topology is crucial, e.g. dressing.\\
Others have relied on \textit{visual} and \textit{visuo-spatial} feature recognition by training deep neural networks to encode RGB or RGB-D observations in a constrained latent space \cite{yan_learning_2020}, \cite{hoque_visuospatial_2021}. By using a full view of the object as input, these methods could address more complex manipulation tasks. However, learning these accurate and compressed latent representations for complex deformable objects requires a quantity of manipulation data that is presently unavailable.\\
Despite the technical limitations of inferring a relevant state representation from real-world observations, several works have proven,  in simulation, the usefulness of extensive and reliable state information to achieve skills such as dressing or dynamic manipulation \cite{clegg_learning_2018}, \cite{9981376}, \cite{10414107}. However, during training, these methods rely on information unavailable in real-world scenarios. Hence, they are either non-transferable or rely on perception transfer from simulated to real-world observations. The generation of perceptually convincing images for simulation is constrained by: 1) most simulated environments used for policy learning offer non-photorealistic rendering capabilities. 2) manually designing assets for high-fidelity rendering of real-life objects is a costly process requiring extensive human work.

As a result, generating high-quality images of simulated data forms a bottleneck in developing generalisable, data-driven, advanced skills for complex textile items' manipulation. In this work, we address this problem by proposing GARField (Garment Attached Radiance Field), a novel, learnable rendering pipeline that allows re-posting deformable meshes for realistic visual data generation from simulation states.
Our contribution is three-fold:
\begin{itemize}
    \item An active perception pipeline enabling realistic data generation from simulation data
    \item A novel relative position coordinate system for efficient learning of fields defined on triangle meshes
    \item A simple training strategy for learning robust geometric and visual features from a limited number of static views
\end{itemize}

In this article, we will first present related works in \ref{sec:rel_w}, then formally pose the problem addressed and our method in \ref{sec:met}, and finally present our experiments, results and a short ablation study in \ref{sec:exp} before concluding.

\section{RELATED WORKS}
\label{sec:rel_w}

\subsection{Differentiable Rendering}
With its impressive rendering accuracy, NeRF \cite{mildenhall2021nerf} has been a cornerstone in establishing differentiable rendering as a relevant tool for multi-view scene capture. In parallel, DeepSDF \cite{Park2019DeepSDFLC} has demonstrated the ability of deep neural architectures to accurately store an object's geometry by fitting over point clouds. By combining these features, NeuS \cite{wang2021neus} constrained NeRF's underlying geometry representation to fit the definition of the signed distance function, thus enabling high-quality capture of both a scene's geometry and appearance. While these methods mainly focused on rendering, they made it possible to integrate information from multiple points of view into a human-interpretable Euclidean 3D space. This architecture allows gradient flow to differentiable scene-embedded functions.\\
Object-NeRF \cite{yang2021objectnerf}, learnt distinct and composable object rendering functions for scene editing. PAC-NeRF \cite{Li2023PACNeRFPA}, PPR \cite{yang2023ppr}, and NDR \cite{Cai2022NDR} paired differentiable rendering with physics-inspired deformation models producing coherent reconstruction over ordered sequences of images.\\
Differentiable rendering has also made its way into the robotics community, where it has been used for grasp detection in Grasp-NeRF \cite{Dai2023GraspNeRF} and Dex-NeRF \cite{IchnowskiAvigal2021DexNeRF}. But also as a 3D \textit{decoder} in \cite{Li20213DNS} for learning a predictive model for visuomotor control during fluid manipulation.\\
Finally, a similar approach has been used recently for data generation in the bin-picking context with COV-NeRF \cite{mishra2024closing}. COV-NeRF introduces an object-composable NeRF that learns pickable objects' geometrical and visual characteristics. Later, the objects are artificially re-arranged randomly to produce new and automatically labelled visuospatial data for downstream perception tasks. Our approach is similar in the general structure of the generation pipeline; however, because of the deformable nature of garments and the complex interaction they can have with their environment, we have relied on stronger shape priors by using triangle meshes, implying significant divergence between the methods.

\subsection{Visual sim-to-real gap in cloth manipulation}
While the gap between simulated data and real-world observation is a problem encountered in all simulation-based robotics, it poses an even greater challenge in textile manipulation as cloth simulators generally produce inaccurate results unless slow and compute-intensive methods, such as FEM, are used \cite{blancomulero2024benchmarking}. The visual sim-to-real gap is a substantial problem resulting in significant performance degradation when transferring from simulation to real-world data, as exemplified by \cite{8967827} or \cite{wu_learning_2023}. The main problem underlying this gap is the challenge of state estimation in a crumpled configuration.\\
Some works have tried to address this problem using shape reconstruction \cite{chi2021garmentnets}, \cite{MEDOR} from half-crumpled observations. They have demonstrated encouraging performance for state estimation.\\
In \cite{SCRACT}, the authors addressed the problem from the data generation angle by producing a dataset of observations with high-quality annotations. This is similar to the approach in this work, but we attain the same goal by automatic generation of convincing observations from known states rather than state estimation from observations with known action sequences.\\
Outside of robotic manipulation, realistic renderings of garments have been tackled in the virtual try-on community \cite{bhatnagar2019mgn}, \cite{corona2021smplicit} with more recent works producing outstanding results \cite{10147324}. Nonetheless, the challenge of generating realistic garment images in the context of try-ons is simplified by the human body constraining the garment shape in a low-occlusion configuration. To our knowledge, our work is the first to present a learnt automatic data generation pipeline for garments in any state.

\section{METHOD}
\label{sec:met}

\begin{figure*}[ht]
  \centering
	\includegraphics[width=\linewidth]{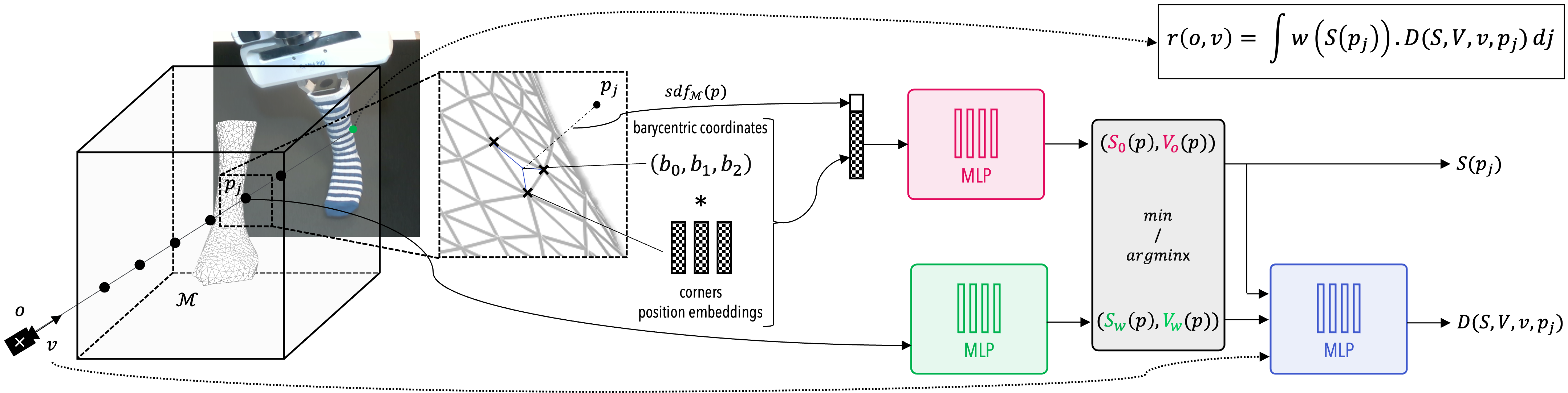}
    \hfill
	\caption{\small GARField models the scene as a composition of signed distance and visual feature fields. The background field is defined in the scene's global coordinates frame. The other fields are \textit{attached} to objects' meshes and can be re-posed. The \textit{mesh-attached} coordinates system projects query points in a coordinate system made up of the point's distance to the mesh's surface and coordinates of the surface-projection of the query point in a bespoke coordinate system built around Laplacian-based position embeddings and barycentric coordinates.}
	\label{fig:architecture}
\end{figure*}

To address the problem of realistic visual data generation, we first develop a learnable composable model of the 3D scene. We then propose a two-staged approach: \textbf{scene capture} and \textbf{scene rendering}.\\
During \textbf{scene capture}, the deformable object is set in a rigid state by stretching it over a solid shape template with minimal self-occlusion. In this configuration, a small set of observations is collected. This limited dataset is then used to train our composable scene model.\\
During the \textbf{scene rendering} stage, we use a deformable object simulator to re-pose the object mesh in an arbitrary number of poses; we then use our model to render novel observations from our learnt visual features.

\subsection{COMPOSABLE SCENE RENDERING}
Given a bounded scene $\mathcal{B} \subset \mathbb{R}^3$ and a deformable garment represented by a triangle mesh $\mathcal{M}$ in an arbitrary state $s=(x, e)$  with $x$ the node positions and $e$ the mesh edges, we aim at rendering the colour $\mathcal{I}_s^c$ and depth $\mathcal{D}_s^c$ images observed from $c$ with known extrinsic and intrinsic matrices.

Similar to \cite{wang2021neus}, we sub-divide the problem to represent the scene geometry on the one hand and, on the other hand, a colour function that attributes a colour to each point in the scene. More precisely, as shown in Fig. \ref{fig:architecture}, we model the scene implicitly through three functions:
\begin{itemize}
    \item $\mathrm{S}: \mathcal{B}\longrightarrow \mathbb{R}$ the signed distance field generated by our scene's geometry
    \item $\mathrm{V}: \mathcal{B} \longrightarrow \mathbb{R}^k$ a feature field associating with each point of $\mathcal{B}$ a latent feature representing its visual aspect
    \item $\mathrm{D}: \mathbb{R}^k \times \mathcal{S}\mathcal{O}(3) \times \mathbb{R} \longrightarrow (0,1)^3$ a decoding function, decoding latent visual features into colour values in the RGB space depending on the viewing direction and signed distance
\end{itemize}

To separate object-centric representations from the background scene, we further subdivide the scene geometry and latent features field by defining $\mathrm{S}_o: \mathcal{B} \longrightarrow \mathbb{R}$ and $\mathrm{S}_w: \mathcal{B} \longrightarrow \mathbb{R}$ the objects and background scene's signed distance function and $\mathrm{V}_o: \mathcal{B} \longrightarrow \mathbb{R}^k$ and the corresponding feature fields: $\mathrm{V}_w: \mathcal{B} \longrightarrow \mathbb{R}^k$.\\
The scene composition is done by combining the signed distance field as:

\begin{equation} \label{eq3}
\begin{split}
    \forall p \in \mathcal{B}, \quad  \mathrm{S}(p) & = \quad \min(\mathrm{S}_w(p), \mathrm{S}_o(p, s))\\
    \mathrm{V}(p) & = \argmin_{(\mathrm{S}_w(p), \mathrm{S}_o(p, s))}(\mathrm{V}_w(p), \mathrm{V}_o(p, s))
\end{split}
\end{equation}

We parametrise $\mathrm{D}$ with an MLP of 4 layers of hidden dimension 256, and $\mathrm{V}_w$ and $\mathrm{S}_w$ with an MLP of 8 layers of hidden dimension 256 and a skip connection at layer 4 . $\mathrm{S}_o(p, s)$, $\mathrm{V}_o(p, s)$ follow a custom architecture described below.\\

Rendering images from our scene model is done by ray shooting through $\mathcal{B}$. For each pixel, the corresponding ray is defined as the set $\{\mathbf{p}(j) = \mathbf{o} + j\mathbf{v}|j \geq 0\}$, where $\mathbf{o}$ is the camera centre, $\mathbf{v}$ is the unit direction of the ray, and $j$ is the distance from the camera centre. For each pixel of $\mathcal{I}_s^c$, the colour is then calculated by integrating along the ray:

\begin{equation} \label{eq1}
\begin{split}
\mathbf{r}(\mathbf{o}, \mathbf{v}) & = \int_0^{+\infty} w_S(j) \cdot r_{S, V}(j) dj\\
w_S(j) &= w(\mathrm{S}(\mathbf{p}(j)) \\
r_{S, V}(j) &= D(\mathrm{V}(\mathbf{p}(j)), \mathbf{v},\mathrm{S}(\mathbf{p}(j)) \\
\end{split}
\end{equation}

And trivially, for each pixel of $\mathcal{D}_s^c$, the depth value is  computed with:
\begin{equation} \label{eq2}
\mathbf{d}(\mathbf{o}, \mathbf{v}) = \int_0^{+\infty} w_S(j) \cdot \mathrm{S}(\mathbf{p}(j)) dj
\end{equation}

With $w$ as the weighting function derived from the signed distance field of the scene from \cite{wang2021neus}.\\

\subsection{MESH-BASED FIELDS}
The key challenge in defining mesh-based fields is defining a surface-attached coordinate system, as garments generally present complex, non-euclidean geometries.\\
Since we suppose access to our object's state as a posed triangle mesh $\mathcal{M}$, we can operate the following decomposition:

\begin{equation} \label{eq4}
  \begin{split}
\forall p \in \mathbb{R}^3, p &= sdf_{\mathcal{M}}(p) \cdot n + q\\
\mathrm{With:} \quad q &= \argmin_{u \in \mathcal{M}}(||p - u||_2)\\
    n &= \frac{p - q}{||p - q||_2}
  \end{split}
\end{equation}

By assuming $\mathcal{M}$ to be watertight, we can guarantee that $sdf_{\mathcal{M}}$ is continuous in $\mathcal{B}$.\\

In practice, we compute the signed distance function $sdf_{\mathcal{M}}$ thanks to the libigl implementation of \cite{1407857}. However, we learn residual geometry that the mesh would not correctly represent by approximating the \textit{true} signed distance function $\mathrm{S}_o$ with a residual MLP of 8 layers with hidden dimension 768 and a skip connection at layer 4.\\

\begin{figure}[ht]
  \centering
  \subfloat[]{\includegraphics[width=0.28\linewidth]{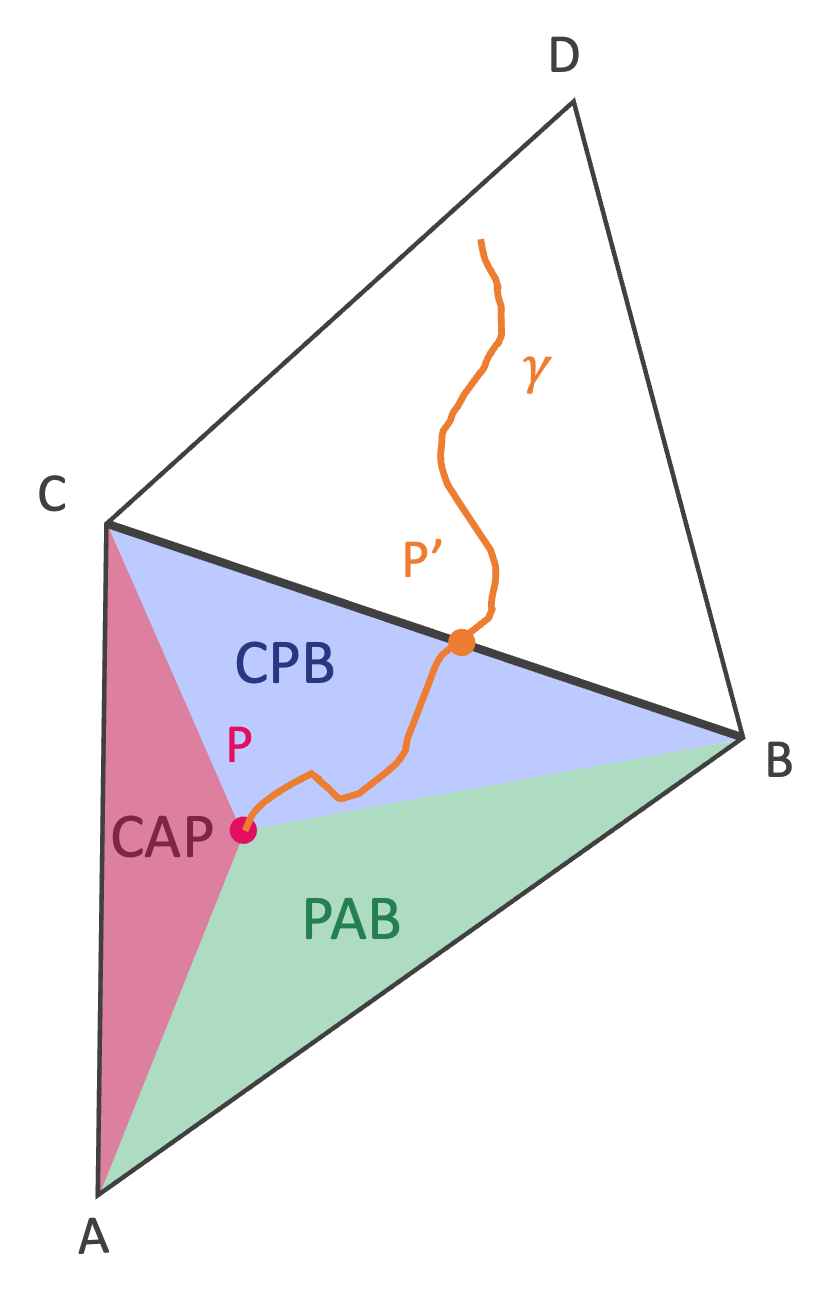}\label{fig:bct_coord:f1}}
  \hfill
  \subfloat[]{\includegraphics[width=0.72\linewidth]{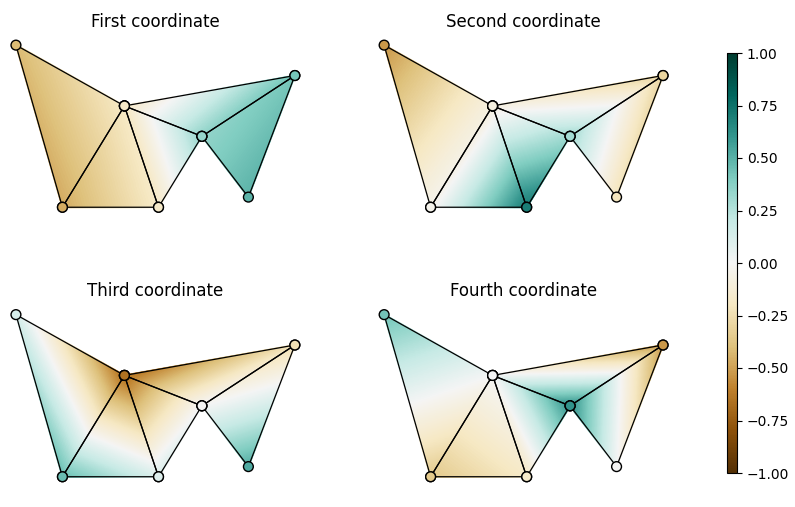}\label{fig:bct_coord:f2}}
  \caption{\small \textbf{(a)} the barycentric coordinates are defined as the ratio of the coloured areas to the area of $ABC$. The $\gamma$ path illustrates over-edge continuity. \textbf{(b)} First coordinates of our system illustrated over an arbitrary mesh.}
\end{figure}

The challenge then becomes defining a continuous coordinate system on the surface of the mesh. A naive approach could be to define a 2D Cartesian coordinate system attached to each triangle or define a point's position in a face with its distance to two or three of the face's corners. However, these coordinate systems would not be over-edge continuous. We define over-edge continuity as the continuity of a surface value at the edge between two connected triangles. A more popular technique, known as UV mapping, is to devise a projection of the mesh surface to a 2D plane and attribute to each node its coordinates in this plane. However, because of the garments' complex geometries, using such techniques produces maps with varying resolution levels over the surface and with a break in continuity along chosen cut-edges.\\
Instead, we opted for an approach based on positional embeddings and barycentric coordinates. Given a triangle $ABC$, as illustrated by Fig. \ref{fig:bct_coord:f1}, and a point $P$ in the triangle, the barycentric coordinates $(b_0, b_1, b_2)$ of $P$ in $ABC$ are defined as follows:

\begin{equation} \label{eq5}
  \begin{split}
  (b_0, b_1, b_2)_{ABC} = (\frac{|\Delta PAB|}{|\Delta ABC|}, \frac{|\Delta CPB|}{|\Delta ABC|}, \frac{|\Delta CAP|}{|\Delta ABC|})
  \end{split}
\end{equation}
with $|\Delta XYZ|$ the area of the triangle $XYZ$.\\
If we define, for each node, an embedding $e_i$ of a chosen dimension $k$ as a unique \textit{code} representing this node's position on the surface, we can then define the coordinate of $P$ in $ABC$ as:
\begin{equation} \label{eq6}
  \begin{split}
  b_0*e_0 + b_1*e_1 + b_2*e_2
  \end{split}
\end{equation}
We can notice that this definition attributes to each node its own embedding. Additionally, given $\gamma$, a path moving over the surface from $ABC$ to $BDC$, and $P'$, the intersection point between $\gamma$ and $BC$. The coordinates of $P'$ approaching from $ABC$ are:
\begin{equation} \label{eq7}
  \begin{split}
  p'_{ABC} & = b_0*e_0 + b_1*e_1 + b_2*e_2\\
  & = 0*e_0 + \frac{|P'B|}{|BC|}*e_1 + \frac{|CP'|}{|BC|}*e_2
  \end{split}
\end{equation}
And from $BDC$:
\begin{equation} \label{eq8}
  \begin{split}
  p'_{BDC} & = b_1*e_1 + b_2*e_2 + b_3*e_3\\
  & = \frac{|P'B|}{|BC|}*e_1 + \frac{|CP'|}{|BC|}*e_2 + 0*e_3\\
  & = P'_{ABC}
  \end{split}
\end{equation}
Thus, we have defined a continuous coordinate system over $\mathcal{M}$.\\

Regarding the choice of the positional embeddings, we opted for the Laplacian-based positional embedding (LPE) introduced by \cite{6789755}. Originating from the field of spectral graph theory, LPE can be defined as a set of embeddings $X = \{X_i \in \mathbb{R}^k; i\in \{1...n\}\}$ that minimises the following minimisation problem:
\begin{equation} \label{eq9}
  \begin{split}
  \min_{X \in \mathbb{R}^{n \times k}} \sum_{ij} a_{ij}||X_i - X_j||_2^2 &= \min_X tr(X^t LX)\\
  s.t. \quad X^tDX &= I_k
  \end{split}
\end{equation}
With $A = (a_{ij})_{(i,j)\in \{1...n\}^2}$ the adjacency matrix of the graph, $D = diag(A)$ and $L = D - A$ the graph Laplacian.\\
As a result, this embedding preserves the notion of proximity between nodes. Additionally, it carries over meshes many of the intuitions of the more familiar sine-cosine embedding used in Euclidian spaces. For instance, it acts as a frequency separation operator on the mesh's surface, allowing for faster learning of our feature fields. This behaviour can be observed in Fig. \ref{fig:bct_coord:f2}, where the first coordinate displays two local optima, the second and third have three, and the fourth has four, thus increasing the amount of variation. Finally, LPE's coordinates values are in the $[-1, 1]$ interval, providing an adequately normalised input to downstream neural networks. During our training, we used an embedding size of $k=256$.\\

\subsection{TRAINING}

\begin{figure*}
  \centering
  \subfloat[]{\includegraphics[width=\linewidth]{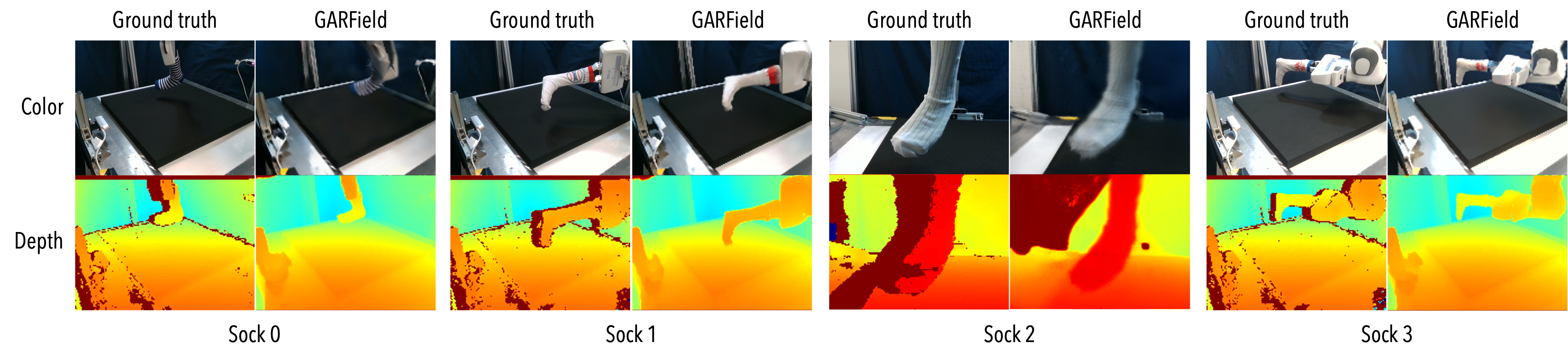}\label{fig:qual_0}}
  \hfill
  \subfloat[]{\includegraphics[width=\linewidth]{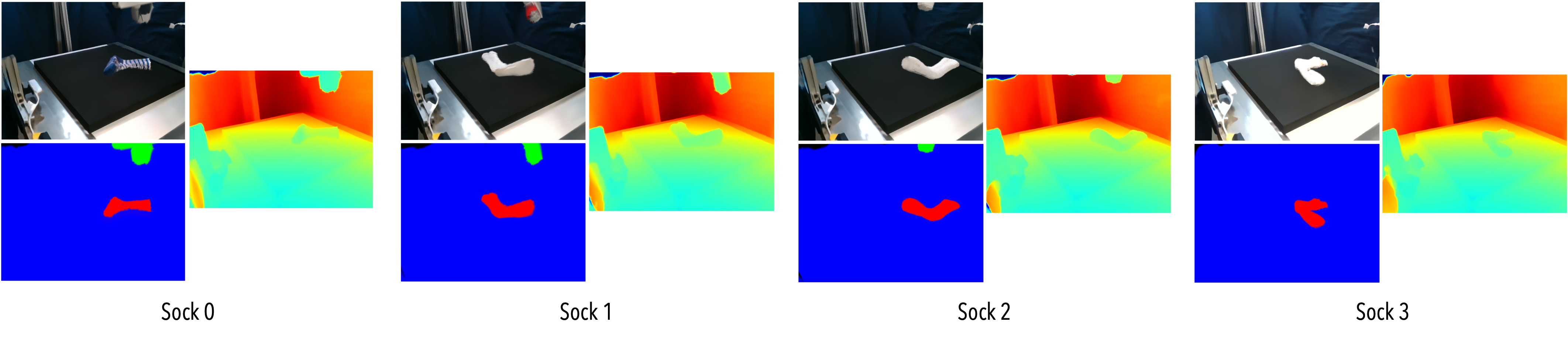}\label{fig:qual_reposed}}
  \caption{\small \textbf{(a)} Qualitative results for reconstruction of training images. \textbf{(b)} Qualitative results for re-posed meshes rendering. Top left: colour, bottom left: masks, right: depth. While the mesh position is more rigid that and usual cotton sock would be, we can observe alignment of geometry and colour features. Additionally, the quality of depth images and masks is on par with training data.}
  \label{fig:qual_results}
\end{figure*}

\textbf{Training loss}\\
We train our architecture by gradient descent optimising with the following loss function:
\begin{equation} \label{eq10}
  \begin{split}
  \mathcal{L} = \alpha \mathcal{L}_{colour} + \beta \mathcal{L}_{depth} + \gamma \mathcal{L}_{reg} + \delta \mathcal{L}_{mesh}
  \end{split}
\end{equation}
With $(\alpha, \beta, \gamma, \delta)$ a set of hyper-parameters, each balancing the influence of each loss term.\\
The colour loss $\mathcal{L}_{colour}$ is defined as an L1-loss. Similarly, the depth loss constrains all points from the depth image $\mathcal{D}_s^c$ to have a 0 signed distance value with an L1-loss:
\begin{equation} \label{eq11}
  \begin{split}
    \mathcal{L}_{depth} = \sum_{p \in \mathcal{D}_s^c} ||\mathrm{S}(p)||_2
  \end{split}
\end{equation}
The Eikonal loss term $\mathcal{L}_{reg}$ is introduced as regularisation term with:
\begin{equation} \label{eq12}
  \begin{split}
    \mathcal{L}_{reg} = \sum_{x} (||\nabla \mathrm{S}(x)||_2 - 1)^2
  \end{split}
\end{equation}
The Eikonal loss is a common component in differentiable rendering, which constrains the norm of the gradient of the signed distance function to tend towards one. This can intuitively be understood since, by definition, moving a distance of one unit in the direction of the gradient of the signed distance function increases the signed distance value by one.\\
Finally, we introduce the $\mathcal{L}_{mesh}$ loss term to constrain $\mathrm{S}_o$ to tend towards ${sdf}_{\mathcal{M}}$:
\begin{equation} \label{eq13}
  \begin{split}
    \mathcal{L}_{mesh} = \sum_{x} ||\mathrm{S}_o(x) - {sdf}_{\mathcal{M}}(x)||_2
  \end{split}
\end{equation}

This loss term is deactivated after ten thousand training iterations by setting $\delta = 0$ to allow $\mathrm{S}_o$ to learn residual real-world geometry that is not modelled by $\mathcal{M}$.\\

\textbf{View direction augmentation}\\
View direction is a necessary input to the colour rendering of the scene as it allows our model to learn how the light from the static light source bounces on surfaces. However, as we use a small number of static cameras in our scene, the network can use the view direction to over-fit each camera and bias its output to fit each camera individually instead of producing a cohesive general scene representation. This can, for instance, lead to poor generalisation capabilities for novel views or to learning a strongly view-dependent feature field. To mitigate this effect, we employ a view direction augmentation by randomly replacing the view direction with a random sample from $\mathcal{S}\mathcal{O}(3)$, during training. In our training, we used a probability of 0.3 for replacement.

\section{EXPERIMENTS}
\label{sec:exp}

\begin{figure}
  \centering
  \subfloat[]{\includegraphics[width=0.5\linewidth]{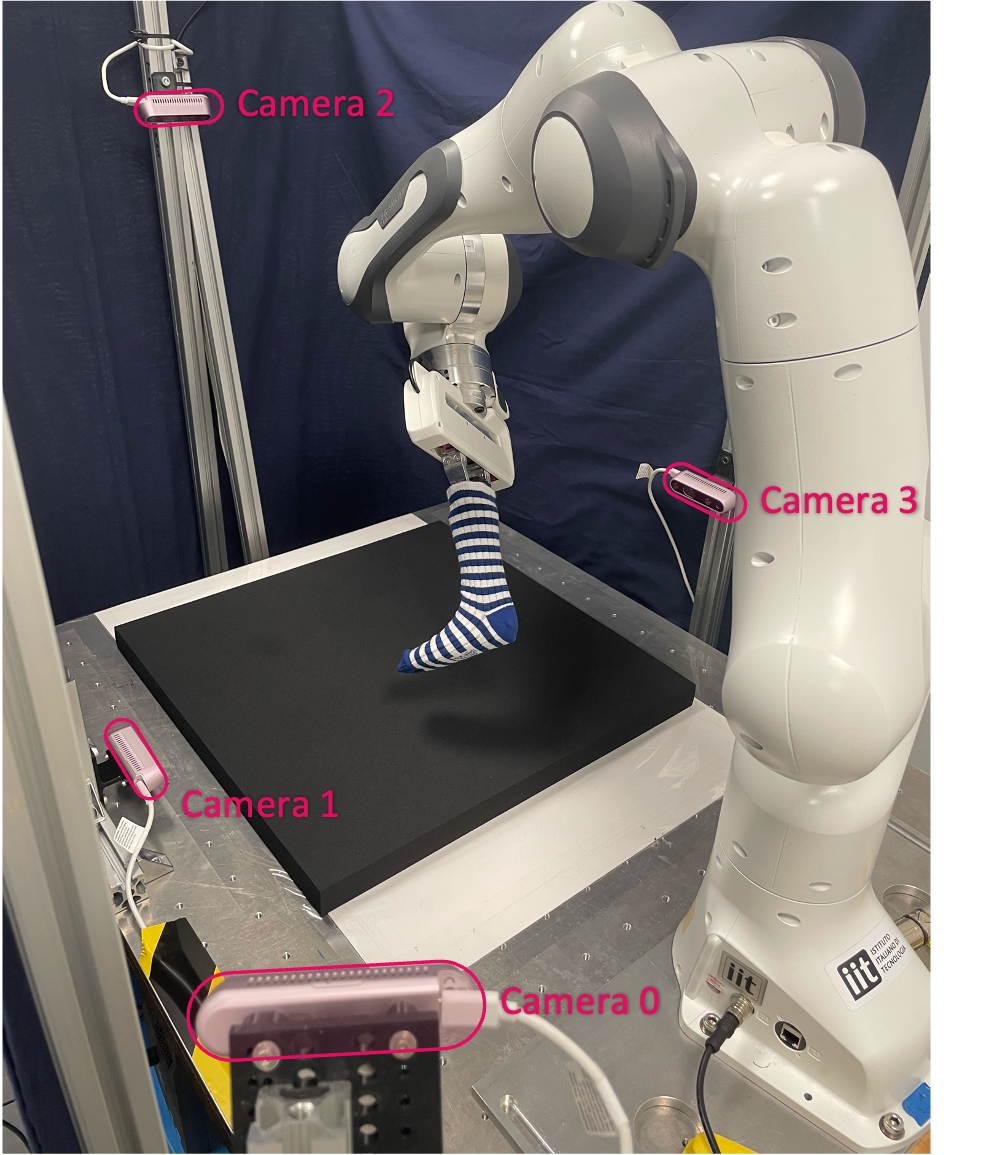}\label{fig:exp_setup}}
  \hfill
  \subfloat[]{\includegraphics[width=0.5\linewidth]{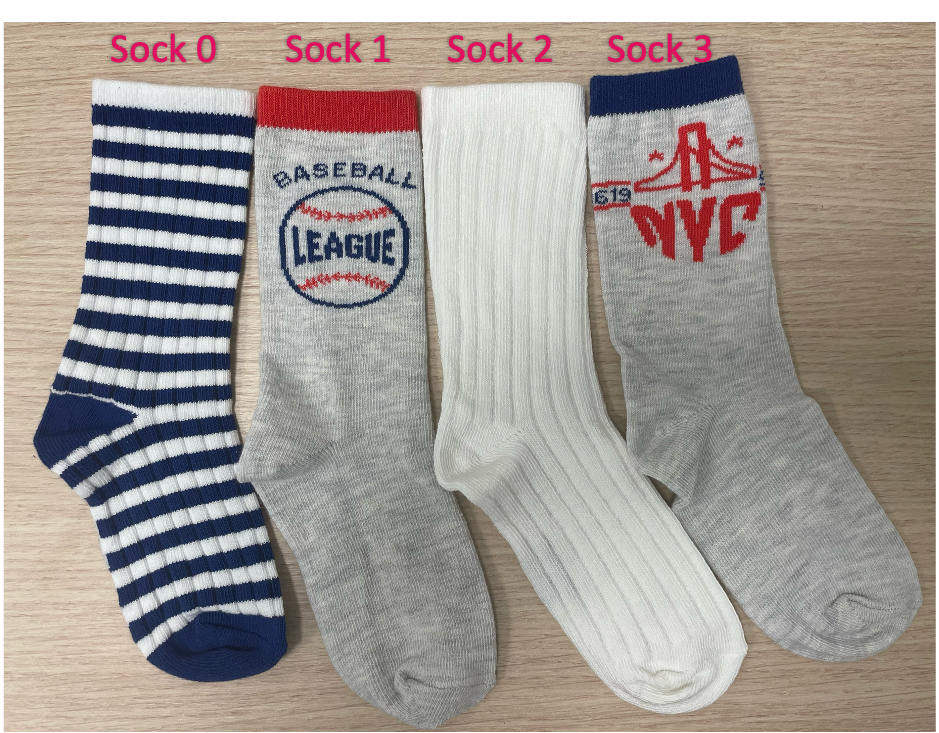}\label{fig:socks}}
  \caption{\small \textbf{(a)} During the capture phase, the Franka-Emika Panda arm poses the shape template in random poses to expose all angles of the deformable object to the four RealSense cameras. \textbf{(b)} We have selected four socks with different patterns of varying sizes to display the performances of our model.}
\end{figure}

The experimental setup, shown in Fig. \ref{fig:exp_setup}, is made up of a Franka-Emika Panda robotic arm, a set of cotton socks of varying appearances, shown in Fig. \ref{fig:socks}, a 3D printed template of known shape and 4 RGB-D Intel RealSense cameras.\\
We sample 40 random poses for each item for the arm and template. For each pose, the four cameras record one RGB-D image each for a total of 160 RGB-D images. In addition, we recorded the joint positions of the arm to locate the mesh shape template correctly in the modelled scene.\\
Training and inference use one NVIDIA's V100 GPU. All results are obtained after training for 90K iterations, which amounts to roughly 22 hours.\\
We present our results in two sections. Firstly, we evaluate the reconstruction fidelity from scene capture images and illustrate the re-posed capabilities. Secondly, we assess the effect of our positional embeddings and view direction augmentation through an ablation study.

\subsection{Rendering}
\begin{figure}
  \centerline{\includegraphics[width=0.8\linewidth]{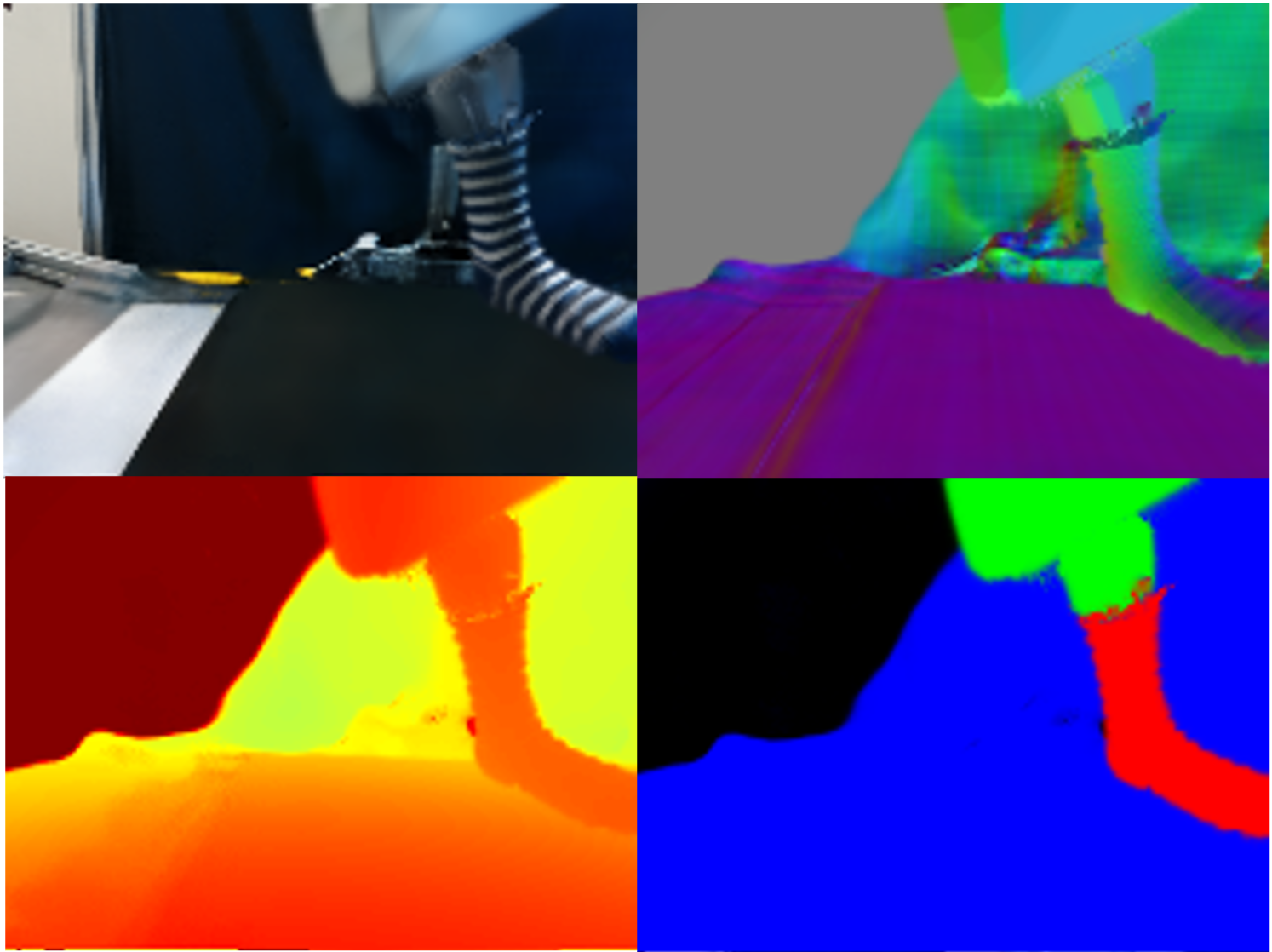}}
  \caption{\small Corresponding outputs of GARField. Top left: colour. Top right: normals. Bottom left: depth. Bottom right: segmentation masks. Black zone in the mask image corresponds to areas outside of the sphere of radius 1 centered on the scene.}
  \label{fig:outs}
\end{figure}

\begin{table}
\caption{\small Image reconstruction performances for each sock}
\begin{center}
\begin{tabular}{ |p{0.1\linewidth}||p{0.23\linewidth}|p{0.23\linewidth}|p{0.23\linewidth}|  }
 \hline
 \textbf{Item}&\textbf{PSNR}&\textbf{SSIM}&\textbf{Geom. error}\\
 \hline
 Sock 0   & $28.14 \pm 2.52$ & $0.82 \pm 0.06$ & $0.09 \pm 0.04$\\
 \hline
 Sock 1   & $25.48 \pm 7.25$ & $0.76 \pm 0.18$ & $0.07 \pm 0.03$\\
 \hline
 Sock 2   & $28.80 \pm 3.06$ & $0.83 \pm 0.06$ & $0.06 \pm 0.02$\\
 \hline
 Sock 3   & $27.48 \pm 3.83$ & $0.83 \pm 0.08$ & $0.06 \pm 0.02$\\
 \hline
\end{tabular}
\label{tab:res_tab}
\end{center}
\end{table}
The performances in image reconstruction are summed up in table \ref{tab:res_tab}, and we present samples of training data reconstructions in Fig. \ref{fig:qual_0} for each sock. The metrics used for comparison are the Peak Signal to Noise Ratio (PSNR) and the Structural Similarity Index Measure (SSIM), as they are common metrics for evaluating the fidelity of reconstructed images. Additionally, we report the geometric error to convey the quality of the reconstructed scene's geometry. The geometric error is defined as the mean of the absolute value of the signed distance function's value computed on surface sample points obtained from depth images. Higher values of PSNR and SSIM demonstrate a higher reconstruction quality, while a lower geometric error shows better geometry reconstruction, as surface points should have a signed distance of zero. All results are reported as the mean and standard deviation of the metric over 40 random images from the sock 0 dataset.\\
We observe from the low value of the error loss in table \ref{tab:res_tab} that our models correctly capture the scene's geometry. Thanks to the integration of data over multiple views and regularity constraints, our architecture is able to produce depth images that are less noisy than the original training images. We can also observe that our architecture is able to capture fine colour details such as the blue stripes on sock 0, the ripples on sock 2, or the red text on the ankle on sock 3. However, we can see that the design with thinner lines in sock 1 has not been correctly captured, as reflected by the lower PSNR and SSIM scores for this item.\\
Additionally to the colour and depth images, GARField trivially produces segmentation masks and surface normals as illustrated in Fig. \ref{fig:outs}. These outputs can be used for downstream tasks such as grasp prediction or segmentation.\\

In Fig. \ref{fig:qual_reposed}, we present the results of rendering re-posed meshes with various amounts of rigidity in the material. The rendering of full-resolution images (640×480px) takes roughly 20 minutes per image.\\
Although the underlying poses generated by the simulator might seem unnatural for cotton socks, the depth images produced by GARField show a correct representation of the geometry of the re-posed socks. The distinctive colour features, such as the blue stripes, are correctly placed on the item. However, we can notice both in reconstructing the training images and in generating the re-posed images that there is some blurring of the visual feature of the socks, and this erases the smaller details, such as the size information present on the bottom of the sole.\\

\subsection{Ablation study}

\begin{figure}
  \centerline{\includegraphics[width=0.8\linewidth]{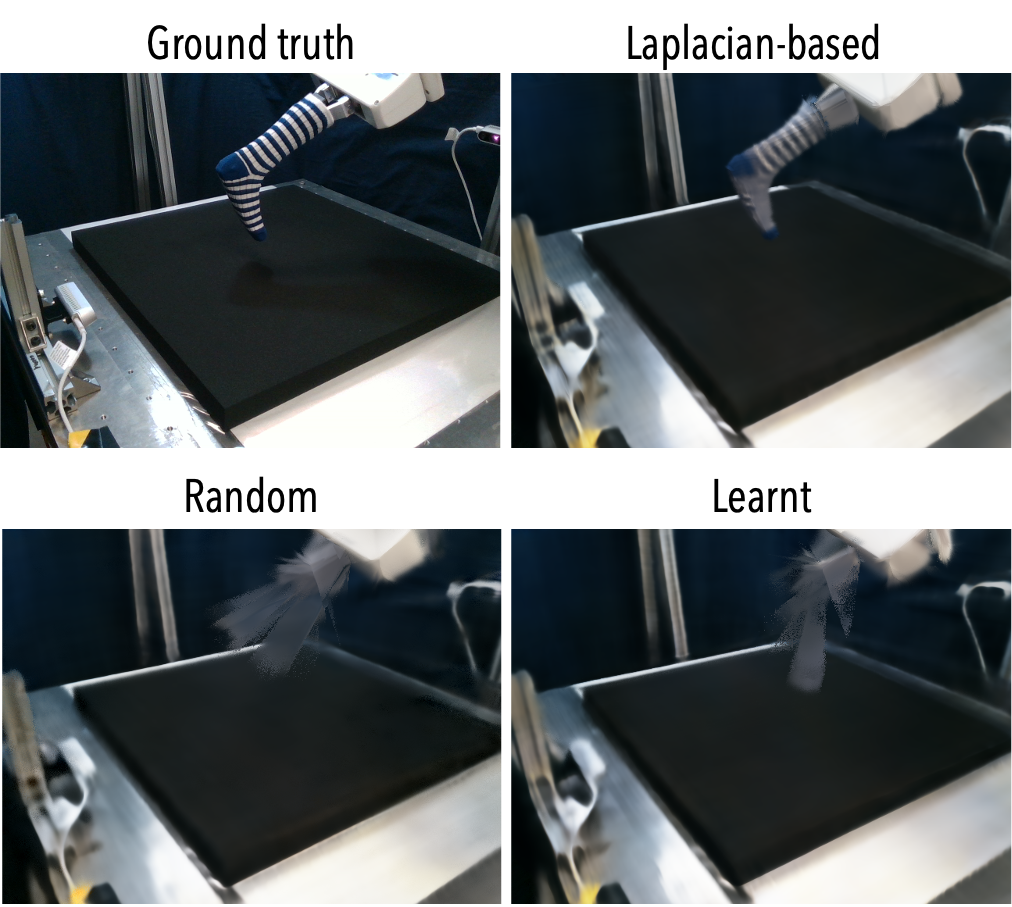}}
  \caption{\small Sampled reconstructed images for training with each positional embedding}
  \label{fig:qual_pe}
\end{figure}

\textbf{Positional embeddings}\\
To evaluate the impact of using the Laplacian embeddings, we compare them with random and learned embeddings.
Random embeddings are a set of vectors randomly sampled before training. Learned embeddings are initialised randomly and jointly optimised with the rendering model.\\
Each positional embedding is evaluated by training the same network architecture for 90K iterations. The metrics are computed on 40 random images of the training dataset for sock 0 (see Fig. \ref{fig:socks}). We report the average value and standard deviation over the dataset in table \ref{tab:pe_tab}. These results confirm that the Laplacian-based position embeddings outperform the other options. Looking at the qualitative data in Fig. \ref{fig:qual_pe}, we notice that the architecture fails to learn the correct geometry and appearance for the non-Laplacian embeddings. This can be attributed to the proximity-preserving nature of Laplacian-based embeddings, which significantly simplifies the task of the MLP in learning a correct surface map.

\begin{table}
\caption{\small Impact of positional embedding on reconstruction}
\begin{center}
\begin{tabular}{ |p{0.2\linewidth}||p{0.2\linewidth}|p{0.2\linewidth}|p{0.2\linewidth}|  }
 \hline
 \textbf{Pos. Emb.}&\textbf{PSNR}&\textbf{SSIM}&\textbf{Geom. error}\\
 \hline
 Laplacian & $\textbf{28.14} \pm 2.52$ & $\textbf{0.82} \pm 0.06$ & $0.09 \pm 0.04$\\
 \hline
 Learnable & $26.04 \pm 2.81$ & $0.81 \pm 0.06$ & $0.08 \pm 0.03$\\
 \hline
 Random & $25.48 \pm 3.78$ & $0.80 \pm 0.05$ & $\textbf{0.07} \pm 0.03$\\
 \hline
\end{tabular}
\label{tab:pe_tab}
\end{center}
\end{table}

\begin{figure}[ht]
  \centerline{\includegraphics[width=\linewidth]{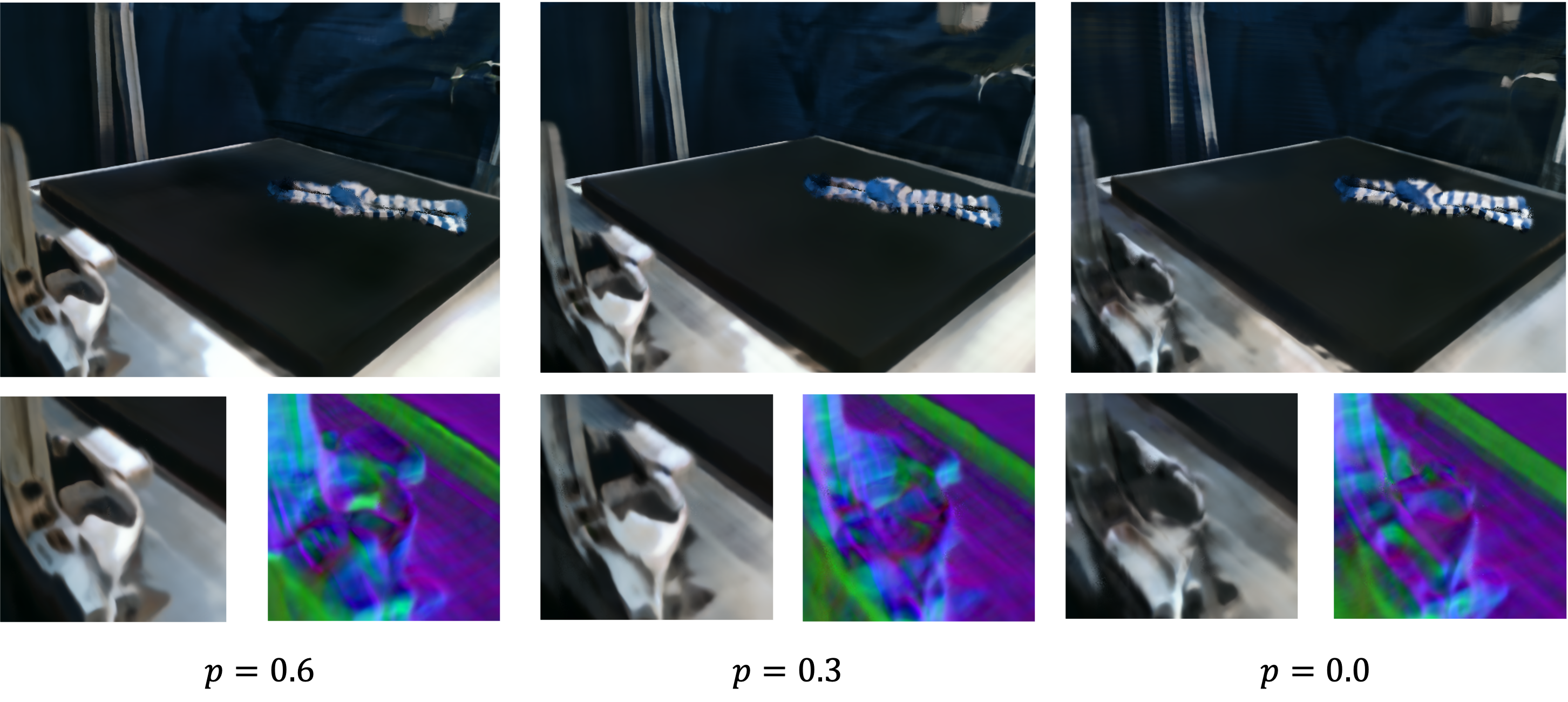}}
  \caption{\small Comparison of details of novel view rendering with different amounts of view direction augmentation during training. \textbf{Top}: Full image rendering. \textbf{Bottom}: Colour and surface normals detail}
  \label{fig:augs}
\end{figure}

\textbf{View direction augmentation}\\
To evaluate the effectiveness of view direction augmentation, we proceed to three trainings on the sock 0 (see Fig. \ref{fig:socks}) dataset with probabilities of 0, 0.3, and 0.6, respectively, for view direction augmentation. All other parameters are kept constant. From the quantitative results presented in table \ref{tab:vd_aug}, we notice a slight degradation in performance in view reconstruction as the augmentation probability grows. However, looking at the sample qualitative result of a novel view rendering presented in Fig. \ref{fig:augs}, we can observe that the decrease in performance is much lower with a higher augmentation percentage, as expected.

\begin{table}
\caption{\small Impact of view direction augmentation}
\begin{center}
\begin{tabular}{ |p{0.2\linewidth}||p{0.2\linewidth}|p{0.2\linewidth}|p{0.2\linewidth}|  }
 \hline
 \textbf{Aug. prob.}&\textbf{PSNR}&\textbf{SSIM}&\textbf{Geom. error}\\
 \hline
 $p=0.0$ & $30.95 \pm 2.32$ & $0.88 \pm 0.04$ & $0.05 \pm 0.02$\\
 \hline
 $p=0.3$ & $28.14 \pm 2.52$ & $0.82 \pm 0.06$ & $0.09 \pm 0.04$\\
 \hline
 $p=0.6$ & $27.65 \pm 3.69$ & $0.82 \pm 0.08$ & $0.07 \pm 0.03$\\
 \hline
\end{tabular}
\label{tab:vd_aug}
\end{center}
\end{table}

\section{CONCLUSIONS}
This paper presents a novel data generation pipeline for garments based on GARField, a fully differentiable neural scene model. For autonomous robots interacting freely with their environment, this tool can enable robots to ``imagine'' how a manipulation plan of a newly discovered object would translate in the observation space. From a more immediate perspective, this tool allows for creating labelled datasets of deformed garments observations from simulated states. This is achieved mainly by exploiting the invariance of visual features in a surface-attached coordinate system.\\
The main limitation of our work is the computational heaviness of the scene rendering and scene capture. Better sampling strategies could be developed to mitigate these challenges.\\
In the future, we would like to further push the fidelity and ease of use of GARField by drawing inspiration, for instance, from the dynamic research community built around NeRF \cite{mildenhall2021nerf}.

\section*{ACKNOWLEDGMENT}

We gratefully acknowledge the Data Science and Computation Facility and its Support Team at Fondazione Istituto Italiano di Tecnologia.\\
The authors gratefully acknowledge the financial support from the Italian  Ministero degli Affari Esteri e della Cooperazione Internazionale (MAECI) as part of the project LEARN-ASSIST (PGR07303).

\addtolength{\textheight}{-10cm}

\bibliographystyle{IEEEtran}
\bibliography{IEEEabrv,references}

\end{document}